\pdfoutput=1

\documentclass[11pt]{article}

\usepackage{acl}

\usepackage{times}
\usepackage{CJKutf8}
\usepackage{latexsym}
\usepackage{xspace}
\usepackage{enumitem}
\usepackage[T1]{fontenc}
\usepackage{marvosym}

\usepackage[utf8]{inputenc}

\usepackage{microtype}

\usepackage{booktabs}
\usepackage{multirow}
\usepackage{arydshln}

\usepackage{xcolor}

\usepackage[ruled,linesnumbered]{algorithm2e}
\SetKwInOut{Parameter}{Parameters}
\usepackage{inconsolata}
\usepackage{graphicx}
\usepackage{dsfont}

\usepackage{adjustbox}
\newcommand{\eg}{\emph{e.g.,}\xspace}
\newcommand{\ie}{\emph{i.e.,}\xspace}
\newcommand{\method}{DUAT}
\newcommand{\dataset}{Challenge-WMT}

\usepackage[most]{tcolorbox}
\definecolor{Ivory2}{HTML}{CFE3E1} 
\definecolor{GreyYellow}{HTML}{E0EBF6}
\definecolor{MyPink}{HTML}{EBDCE1}
\definecolor{MyGreen}{HTML}{DCEBDC}
\definecolor{MyPurple}{HTML}{C7C7F4}
\definecolor{Green2}{HTML}{4FAD5B}
\definecolor{Blue2}{HTML}{2F6EBA}
\definecolor{Purple3}{HTML}{9392BB}
\definecolor{Blue3}{HTML}{D4E6EC}
\definecolor{Green3}{HTML}{D8E3AE}
\definecolor{GreenFig2}{HTML}{4FAD5D}
\definecolor{Blue4}{HTML}{C7E4EA}
\definecolor{Green4}{HTML}{D1DFA1}

\newcommand\Demonstrations[1]{\colorbox{Ivory2}{\raisebox{0pt}[5pt][0pt]{{#1}}}}
\newcommand\SourceSentence[1]{\colorbox{GreyYellow}{\raisebox{0pt}[5pt][0pt]{{#1}}}}
\newcommand\DraftTranslation[1]{\colorbox{MyPink}{\raisebox{0pt}[5pt][0pt]{{#1}}}}
\newcommand\DifficultWords[1]{\colorbox{MyPurple}{\raisebox{0pt}[5pt][0pt]{{#1}}}}
\newcommand\Interpretations[1]{\colorbox{MyGreen}{\raisebox{0pt}[5pt][0pt]{{#1}}}}

\newtcolorbox{mybox}[2][]{
	width=\columnwidth,
	colback = gray!8, 
	colframe = gray!8, 
	boxsep=0pt,left=10pt,right=10pt,top=0pt,bottom=0pt,
	fontupper=\fontsize{9}{10}\selectfont,
	title=#2,#1}
 
\title{Aligning Translation-Specific Understanding to General Understanding in Large Language Models}

\author{Yichong Huang$^{\dag}$, Baohang Li$^{\dag}$, \textbf{Xiaocheng Feng}$^{\dag \ddag}$, Wenshuai Huo$^{\dag \ddag}$, Chengpeng Fu$^{\dag \ddag}$, \\
\textbf{Ting Liu}$^{\dag}$\textbf{, Bing Qin}$^{\dag \ddag \textrm{\Letter}}$\\
  $^{\dag}$Harbin Institute of Technology\quad \quad \quad $^\ddag$ Peng Cheng Laboratory\\
  \texttt{\{ychuang,xcfeng,baohangli,cpfu,wshuo,tliu,qinb\}@ir.hit.edu.cn}
  \\}

\begin{document}
\begin{CJK}{UTF8}{gbsn}
\maketitle
\begin{abstract}
Large Language models (LLMs) have exhibited remarkable abilities in understanding complex texts, offering a promising path towards human-like translation performance.
However, this study reveals the misalignment between the translation-specific understanding and the general understanding inside LLMs.
This understanding misalignment leads to LLMs mistakenly or literally translating some complicated concepts that they accurately comprehend in the general scenarios (\eg QA).
To align the translation-specific understanding to the general one, we propose a novel translation process, \method~(\textbf{\underline{D}}ifficult words \textbf{\underline{U}}nderstanding \textbf{\underline{A}}ligned \textbf{\underline{T}}ranslation), explicitly incorporating the general understanding on the complicated content incurring inconsistent understanding to guide the translation. 
Specifically, \method~performs cross-lingual interpretation for the difficult-to-translate words and enhances the translation with the generated interpretations. 
Furthermore, we reframe the external tools to improve \method~in detecting difficult words and generating helpful interpretations. 
We conduct experiments on the self-constructed benchmark \dataset\footnote{The dataset is available at: \href{https://github.com/OrangeInSouth/ChallengeWMT}{ChallengeWMT}

\quad \textrm{\Letter} means corresponding author.}, consisting of samples that are prone to mistranslation.
Human evaluation results on high-resource and low-resource language pairs indicate that \method~significantly facilitates the understanding alignment, which improves the translation quality (up to +3.85 COMET) and reduces the literality of the translation by - 25\% $\sim$ - 51\%.
\end{abstract}

\section{Introduction} 
Recently, large language models (LLMs) have demonstrated remarkable language understanding and generation, paving the way for a higher level of performance in machine translation~\cite{zhao2023survey,openai2023gpt4,jiang2023mistral,workshop2023bloom}.  
However, existing research reports that LLMs have yet to achieve as significant advances in machine translation as they have achieved in other natural language processing fields~\cite{hendy2023good,pang2024salute,10.5555/3618408.3620130,jiao2023chatgpt,zhu2023extrapolating,lou-etal-2023-cceval}.

\begin{figure}[t]
  \centering
    \includegraphics[clip,width=1.0\columnwidth,]{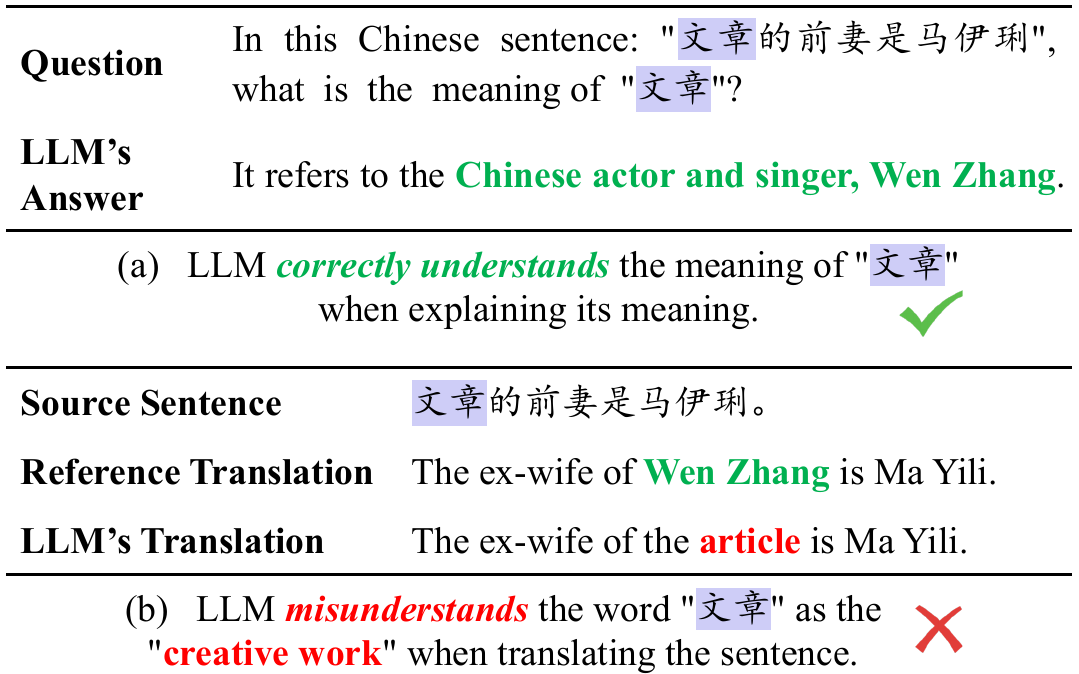}
    \caption{Illustration of the misalignment between the general understanding (Fig a) and the translation-specific language understanding (Fig b) inside the LLM (\texttt{gpt-3.5-turbo-0125}).
    More examples are reported in Appendix~\ref{section:more examples}.
    }
  \label{fig:motivation}
\vspace{-0.3cm}
\end{figure}
In this study, we discover the misalignment between the general understanding and translation-specific understanding inside LLMs, as illustrated in Fig.\ref{fig:motivation}.
This understanding misalignment leads to LLMs mistakenly or literally translating some complicated concepts that they accurately comprehend in general scenarios. We refer to these failures as language models' \textbf{generalization failures} on translation.
Human evaluation on a total of 600 sampled sentences across six language pairs show that generalization failures account for a considerable proportion of all mistranslations (\textbf{16\%-32\%}), indicating serious understanding misalignment (\S\ref{subsection:human evaluation}).


To align the translation-specific understanding to the general one, we propose a novel translation process, \method~(Difficult words Understanding Aligned Translation), explicitly incorporating the general understanding on the complicated content incurring inconsistent understanding to guide the translation. 
Specifically, \textsc{\method} first detects the difficult-to-translate words in the source sentence, which could cover the generalization failures intuitively.
Next, the LLM is prompted to interpret each difficult word with the target language, \ie cross-lingual interpretation, unleashing the powerful general understanding and transforming this understanding into the target language space.
After that, \textsc{\method} conducts translation under the guidance of these interpretations.
Unlike the CoT-based process mimicking junior translators to sequentially translate all words~\cite{peng-etal-2023-towards}, \textsc{\method} works like senior translators 
 to analyze the complicate words, which helps the model deep understand the source sentence and produces more nuanced translations.
Furthermore, we reframe the external tool of token-level QE~\cite{rei-etal-2023-inside} to enhance the detection of difficult words, and design a strategy of interpretation quality control to filter hallucinated interpretations based on sentence-level QE~\cite{rei-etal-2020-comet}.

\begin{figure*}[t]
  \centering
    \includegraphics[clip,width=2.0\columnwidth,]{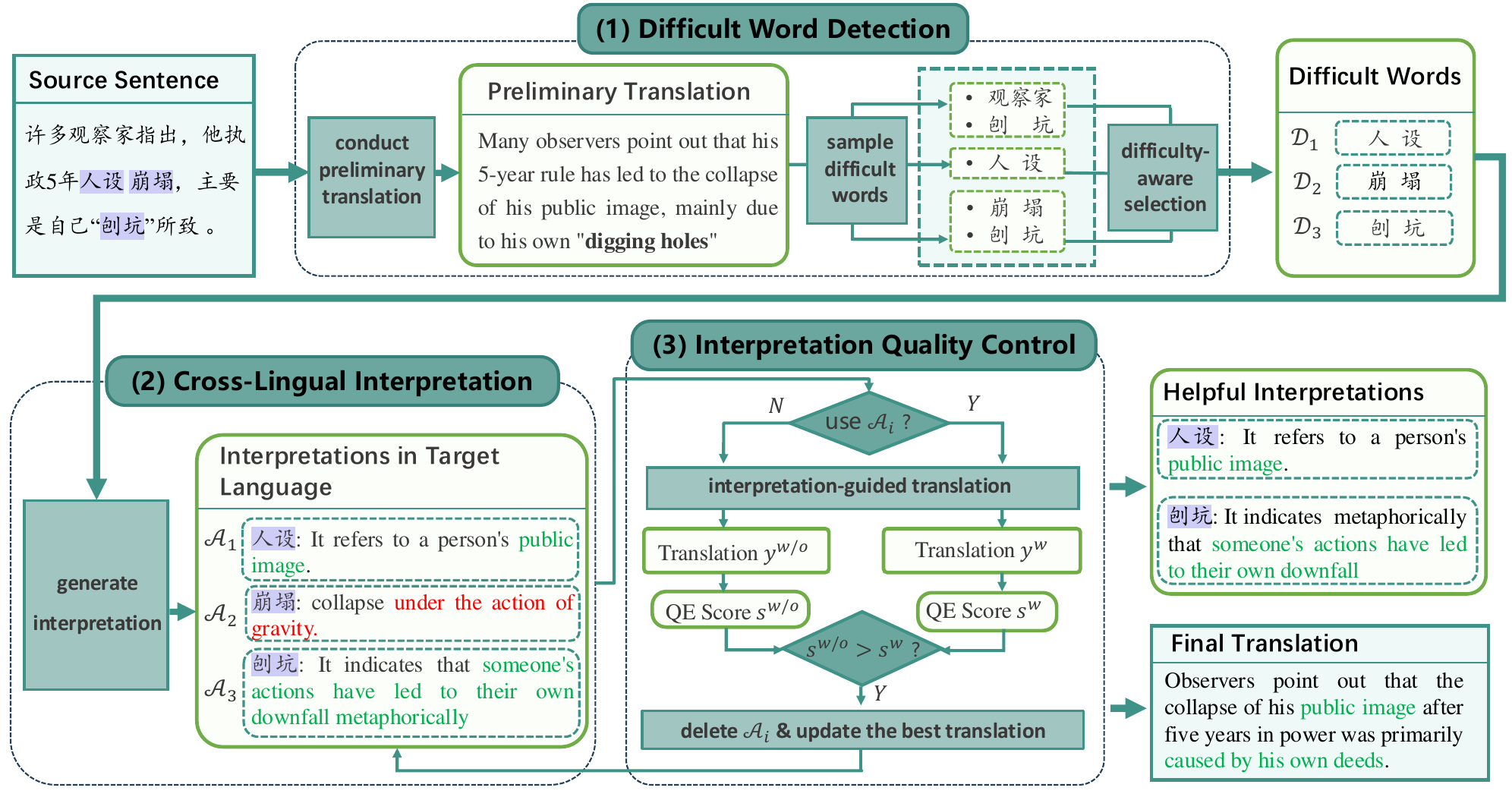}
    \caption{\textsc{\method} framework. The\colorbox{MyPurple}{\raisebox{0pt}[5pt][0pt]{{purple}}} spans indicate the difficult-to-translate words, the \textcolor{GreenFig2}{green} spans indicate the correct translation/interpretation, and the \textcolor{red}{red} spans indicate the incorrect ones.}
  \label{fig:\method}
\end{figure*}

To better analyze the understanding misalignment, we proposed the \dataset~benchmark, which contains more sentences prone to mistranslation. These sentences were collected from multi-year WMT datasets and represent difficult samples that multiple state-of-the-art (SOTA) systems translate poorly. 
Human evaluation results indicate that DUAT significantly facilitates the understanding alignment, reducing 80\%$\sim$88\% of generalization failures.
Moreover, this alignment improves the translation quality, as evidenced by automatic metrics (up to +3.85 COMET), and alleviates translation literalness by -25\% $\sim$ -51\%.


\section{Background} 
\subsection{LLM-based MT}
\label{subsection: LLM-based machine translation}
Considering the translation from source language $L_s$ to target language $L_t$, LLM-based machine translation converts the source sentence $x$ to an instruction using a translation-specific template and generates the translation by feeding the instruction to the LLM $\theta$.
To make the LLM better follow the instruction, the in-context learning (ICL) strategy~\cite{NEURIPS2020_1457c0d6, dong2023survey} injects a few examples/demonstrations of translation into the instruction, which is shown as:
\begin{mybox}
    \ Request: Please translate the $[L_s]$ sentence into $[L_t]$. \\ 
    \textit{\#\, followed by} [\Demonstrations{$N$ Demonstrations $\mathcal{E}^{mt}$ }] \\
    Source Sentence: \ \  [\SourceSentence{Source Sentence $x$}] 
\end{mybox}
Formally, the LLM-based MT generates the translation with ICL as:
\begin{equation}
\label{eq:ICL}
\begin{aligned}
\hat{y} = \mathop{argmax} P_{\theta}(\mathcal{E}^{mt}, x),
\end{aligned}     
\end{equation}
where $\mathcal{E}^{mt}=\{(x^i, y^i)\}_{i=1}^N$ is the demonstrations set of translation.

\subsection{Quality estimation (QE)}
\label{subsection:QE}
QE for machine translation, \ie reference-free MT evaluation, aims to predict the quality of the given translation only according to the source sentence, which has shown auspicious correlations with human judgments~\cite{rei-etal-2020-comet,rei-etal-2021-references}. Given a source sentence $x$ and a
translation $y$, QE score is denoted as $\psi(y\ |\ x )$.

Thanks to the recent advance in the interpretability of neural MT metrics~\cite{rei-etal-2023-inside}, token-level QE is proposed to score the error degree of the given translation span by calculating the misalignment of this span against the source sentence.
Given a source sentence $x$ and the candidate translation $\widetilde{y}$, token-level QE $\phi(\cdot)$ annotates the error degree of the specific span $w^t$ in the translation, \ie $\phi(w^t \ | \ \widetilde{y}, x)$ where $w^t \in \widetilde{y}$.

\section{Approach: \textsc{\method}} 
In this section, we first introduce our translation framework \textsc{\method} (\S\ref{subsection:\method}).
Specifically, \textsc{\method} consists of three components: \textit{difficult word detection} (\S\ref{subsection:detection of difficult words}), \textit{cross-lingual interpretation} (\S\ref{subsection:cross-lingual interpretation}), and \textit{interpretation quality control} (\S\ref{subsection:final translation}).
To make the LLM follow the procedure of each component as expected, we adopt the in-context learning strategy and design an automatic method for constructing demonstrations of \textsc{\method} (\S\ref{subsection:Automatic Construction of Demonstrations}).

\subsection{Framework}\label{subsection:\method}
The progress of \textsc{\method} is illustrated in Fig.\ref{fig:\method}.
Given the source sentence, \textsc{\method} first detects the difficult words or phrases in the source sentence. 
Once the difficult words are identified, \textsc{\method} requests the LLM to interpret each difficult word with the target language, unleashing the powerful understanding capability inside the LLM and transforming these understandings into the target language space.
Finally, to avoid the interference of incorrect and useless interpretations, \textsc{\method} removes the negative interpretations through the interpretation quality control and outputs the final translation guided by the helpful interpretations.

\subsection{Step-1: Difficult Word Detection} \label{subsection:detection of difficult words}
In practice, we found that directly inquiring LLMs to identify difficult-to-translate words is challenging.
To tackle this challenge, we first conduct a preliminary translation for the given source sentence and then extract the \textbf{mistranslated words and phrases} in the source sentence \textbf{as the difficult words}. 
Concretely, we invent \textsc{\method-I} to do this leveraging the \textit{\textbf{I}ntrinsic} ability of LLMs at first.

\paragraph{\textsc{\method-I}.} Given source sentence $x$, \textsc{\method-I} first obtains the preliminary translation $\widetilde{y}$ (also known as \textit{draft translation}) by prompting the LLM to translate $x$ with the in-context learning strategy, which is shown in Eq.~(\ref{eq:ICL}).
Next, the LLM is requested to output the difficult words based on the source sentence and the preliminary translation:

\begin{mybox}
    \ Request: Given a $[L_s]$ sentence and its draft $[L_t]$ translation, output the mistranslated words and phrases in the $[L_s]$ sentence. \\ 
  \textit{\#\, followed by} [\Demonstrations{$N$ Demonstrations $\mathcal{E}^{diff}$}]  \\
  Source Sentence: \ \  [\SourceSentence{Given Sentence $x$}] \\
  Draft Translation:\;  [\DraftTranslation{Draft Translation $\widetilde{y}$}]
\end{mybox}
\textsc{\method-I} obtains the difficult word list $\mathcal{D}$ via performing \textit{greedy decoding} on the LLM:
\begin{equation}
\label{eq:\method-I}
\begin{aligned}
\mathcal{D} = \mathop{argmax} P_{\theta}(\mathcal{E}^{diff}, x, \widetilde{y}\ ),
\end{aligned}     
\end{equation}
where $\theta$ is the LLM, which is prompted with $N$ demonstrations of difficult word detection $\mathcal{E}^{diff} = \{x^i,\widetilde{y}^i, \mathcal{D}^i\}^N_{i=1}$.

\paragraph{\textsc{\method-E}.} It is frequently observed that the LLM fails to recognize the mistranslated words, due to their limitation in self-knowledge~\cite{yin-etal-2023-large}
Therefore, we devise \textsc{\method-E} to boost the detection with the \textit{\textbf{E}xternal} tool. First, \textsc{\method-E} requests the LLM with the same prompt as \textsc{\method-I} while performing \textit{temperature sampling} for $K$ times. Next, the union of all sampling results is taken as the candidate set of difficult words $\mathcal{D}^{cand}$:
\begin{equation}
\label{eq:\method-E}
\begin{aligned}
\mathcal{D}^{cand} = \mathop{\cup}\limits_{k=1}^{K} \mathcal{D}_k \sim P_{\theta}(\mathcal{E}^{diff}, x, \widetilde{y}, T), 
\end{aligned}     
\end{equation}
where $T$ is the hyperparameter of sampling temperature, which is set to $0.5$ to capture more candidates, and $K$ is set to 5 empirically.

Finally, \textsc{\method-E} annotates each candidate word with its degree of misalignment with respect to the draft translation, which reflects the translation-specific difficulty.
To implement this function, we adopt an external tool of token-level QE $\phi(\cdot)$.
As shown in \S\ref{subsection:QE}, token-level QE is originally used to score the mistranslation degree of the given translation span with respect to the source sentence, \ie $\phi(w^t\ |\ \widetilde{y} , x )$ where $w^t \in \widetilde{y}$.
Differently, we utilize this tool in a dual manner. That is, we use $\phi(\cdot)$ to annotate the misalignment degree of the given \textbf{\textit{source}} span with respect to the translation, \ie $\phi(w^s \ |\  x, \widetilde{y}\ )$ where $w^s \in x$.
Formally, the misalignment score of each difficult word candidate is calculated as:
\begin{equation}
\label{eq:misalignment}
\begin{aligned}
\phi(d) = \phi(d\ |\ x, \widetilde{y}\ ), d \in \mathcal{D}^{cand}.
\end{aligned}     
\end{equation}
Then, \textsc{\method-E} selects candidates with misalignment score $\phi(d) > \tau$, where $\tau$ is the hyperparameter named the difficulty threshold.
We refer to this procedure as the \textit{difficulty-aware selection} in Fig.\ref{fig:\method}.

\subsection{Step-2: Cross-Lingual Interpretation} \label{subsection:cross-lingual interpretation}
After the difficult words in the source sentence are detected, \textsc{\method} lets the LLM generate the interpretation of each difficult word via requesting:
\begin{mybox}
    \ Request: Given a $[L_s]$ sentence, provide the concise interpretation for each difficult word with the $[L_t]$. \\ 
  
   \textit{\#\, followed by} [\Demonstrations{$N$ Demonstrations $\mathcal{E}^{intp}$ }] \\
  
  Source Sentence: \ \  [\SourceSentence{Given Sentence $x$}] \\
  Difficult Words: [\DifficultWords{Difficult Words $\mathcal{D}$}]
\end{mybox}
Through access to the LLM, the interpretation set $\mathcal{A}$ is obtained:
\begin{equation}
\label{eq:cross-lingual interpretation}
\begin{aligned}
\mathcal{A} = \mathop{argmax} P_{\theta}(\mathcal{E}^{intp}, x, \mathcal{D}),
\end{aligned}     
\end{equation}
where $\mathcal{E}^{intp} = \{x^i, \mathcal{D}^i, \mathcal{A}^i\}^N_{i=1}$, which is the demonstrations of the cross-lingual interpretation.

\paragraph{Prob and cons.} Under the guidance of generated interpretations, \method~can align the translation-specific understanding to the general one.
However, LLMs may generate hallucinated interpretations sometimes (\textit{e.g.,} the interpretation of "崩塌" in Fig.\ref{fig:\method}), which biases the resulting translation from the original semantics.
Besides, helpless interpretations that can not provide useful information also pose a risk of disturbing the translation process.

\subsection{Step-3: Interpretation Quality Control} \label{subsection:final translation}
To overcome the potential interference of the generated negative interpretations, \method~removes them through the interpretation quality control (\textbf{IQC}) and outputs the final translation guided by the helpful interpretations.

Concretely, given a set of interpretations $\mathcal{A}$, \textsc{\method} ablates each interpretation $\mathcal{A}_i$ sequentially and uses the remaining interpretations to guide the translation. 
The \textbf{interpretation-guided translation} is implemented in a fashion of \textit{refinement}:
\begin{mybox}
    \ Request: Given a $[L_s]$ sentence and its draft $[L_t]$ translation, please revise the translation according to the interpretations of the difficult words.\\ 
  
   \textit{\#\, followed by} [\Demonstrations{$N$ Demonstrations $\mathcal{E}^{igt}$ }] \\
  
    Source Sentence: \ \  [\SourceSentence{Given Sentence $x$}] \\
  Draft Translation:\;  [\DraftTranslation{Draft Translation $\widetilde{y}$}] \\
  Interpretations of Difficult Words: \\ 
  
    [\Interpretations{Interpretations $\mathcal{A}$}]
\end{mybox}
Formally, the translation is obtained as:
\begin{equation}
\label{eq:interpretation-guided translation}
\begin{aligned}
\hat{y} = \mathop{argmax} P_{\theta}(\mathcal{E}^{igt}, x, \widetilde{y}, \mathcal{A}),
\end{aligned}     
\end{equation}
where $\mathcal{E}^{igt} = \{x^i, \widetilde{y}^i, \mathcal{A}^i, \hat{y}^i\}$, which is the demonstration set of interpretation-guide translation.

If the better translation performance is achieved by ablation, which is measured by the QE\footnote{We use wmt21-comet-qe-da as the QE scorer.} tool due to the unavailable access to the reference translation, the interpretation $\mathcal{A}_i$ is removed from $\mathcal{A}$ and the current translation is taken as the best translation.
We also detail this process in Alg.\ref{alg:IQC}.

\subsection{Demonstrations Synthesis for \textsc{\method}} \label{subsection:Automatic Construction of Demonstrations}
To make the LLM follow the procedure of \textsc{\method} as expected, we adopt the ICL strategy.
Common practice constructs demonstrations manually, necessitating human translators proficient in $N \times (N-1)$ language pairs for $N$ languages.
To overcome this considerable cost, we devise a method for synthesizing high-quality demonstrations of \textsc{\method} based on parallel data.
This process of synthesizing demonstrations is accomplished in a manner of post-explanation by asking the LLM to compare the baseline translation and the reference translation.
We describe this process in Appendix~\ref{subsection:details of demonstration synthesis}.

\begin{table*}[!t]
\small
\renewcommand\tabcolsep{1.5pt}
\renewcommand\arraystretch{1.0}
\centering
\begin{tabular}{lcccccccccccccc}

\toprule
\multirow{2}{*}{\textbf{Methods}} & \multicolumn{2}{c}{\textbf{En$\Rightarrow$Zh}} & \multicolumn{2}{c}{\textbf{Zh$\Rightarrow$En}} & \multicolumn{2}{c}{\textbf{En$\Rightarrow$Et}} & \multicolumn{2}{c}{\textbf{Et$\Rightarrow$En}} & \multicolumn{2}{c}{\textbf{En$\Rightarrow$Is}} & \multicolumn{2}{c}{\textbf{Is$\Rightarrow$En}} & \multicolumn{2}{c}{\textbf{Average}}  \\

\cmidrule(lr){2-3}
\cmidrule(lr){4-5}
\cmidrule(lr){6-7}
\cmidrule(lr){8-9}
\cmidrule(lr){10-11}
\cmidrule(lr){12-13}
\cmidrule(lr){14-15}
 & \scriptsize{COMET}   & \scriptsize{BLEURT}   &\scriptsize{COMET}   & \scriptsize{BLEURT}   &\scriptsize{COMET}   & \scriptsize{BLEURT}   & \scriptsize{COMET}   & \scriptsize{BLEURT}   & \scriptsize{COMET}   & \scriptsize{BLEURT}   & \scriptsize{COMET}   & \scriptsize{BLEURT} & \scriptsize{COMET}   & \scriptsize{BLEURT}  \\
\midrule
\multicolumn{15}{c}{\textbf{\textit{Existing Systems}}} \\
\textbf{Google} & 74.85 & 54.95 & 68.21 & 52.65 & 79.11 & 68.71 & 78.83 & 65.46 & 76.17 & 59.67 & 78.70 & 66.54 & 75.98 & 61.33\\
\textbf{NLLB} & 68.77 & 47.77 & 60.09 & 45.14 & 74.20 & 63.13 & 74.35 & 60.87 & 69.37 & 52.08 & 72.55 & 59.66 & 69.89 & 54.78\\
\textbf{GPT-4} & 76.15 & 55.37 & 70.77 & 58.85 & 80.25 & 70.80 & 77.83 & 65.48 & \textbf{77.33} & \textbf{61.15} & \textbf{79.39} & \textbf{68.40} & 76.95 & 63.34\\

\midrule
\multicolumn{15}{c}{\textbf{\textit{Baselines}}} \\
\textbf{Zero-shot} & 74.89 & 54.25 & 71.27 & 58.24 & 80.67 & 69.10 & 74.93 & 61.75 & 71.17 & 53.12 & 76.22 & 64.05 & 74.86 & 60.09\\
\textbf{ICL} & 75.47 & 55.79 & 72.22 & 59.56 & 80.9 & 69.93 & 79.40 & 66.63 & 73.19 & 54.87 & 77.52 & 65.81 & 76.45 & 62.10\\
\textbf{\; +CoT} & 73.85 & 53.42 & 71.35 & 57.90 & 78.03 & 66.03 & 76.78 & 63.97 & 69.72 & 50.98 & 76.55 & 64.54 & 74.38 & 59.47\\
\textbf{\; +Keywords} & 73.93 & 55.10 & 71.22 & 59.18 & 78.63 & 70.01 & 77.79 & 66.30 & 70.33 & 54.31 & 74.55 & 64.40 & 74.41 & 61.55\\
\textbf{\; +Topic} & 75.83 & 53.60 & 72.46 & 57.98 & 80.98 & 66.78 & 79.20 & 64.64 & 72.77 & 51.98 & 76.49 & 62.06 & 76.29 & 59.51\\
\textbf{\; +SimDems} & 75.22 & 55.10 & 72.20 & 59.16 & 81.24 & 70.42 & 79.11 & 66.29 & 72.70 & 54.04 & 76.78 & 64.43 & 76.21 & 61.57\\

\midrule
\multicolumn{15}{c}{\textbf{\textit{Ours}}} \\
\textbf{\method-I} & 76.92 & 56.16 & 72.94 & 59.94 & 82.92 & 72.19 & 79.96 & \textbf{67.05} & 76.64 & 57.71 & 78.45 & 66.80 & 77.97 & 63.31\\
\textbf{\method-E} & \textbf{77.57} & \textbf{56.86} & \textbf{73.25} & \textbf{60.16} & \textbf{83.07} & \textbf{72.30} & \textbf{80.01} & 66.91 & 77.04 & 57.38 & 78.70 & 66.93 & \textbf{78.27} & \textbf{63.42}\\
\bottomrule
\end{tabular}
\caption{Main results on \dataset. 
The bold indicates the highest value.
`+SimDems' represents the translation strategy with demonstrations similar to the source sentence.
The strategies `+Topic', `+Keywords', and `+SimDems' are proposed in MAPS.
The baselines and our approaches are implemented based on \texttt{GPT-3.5-turbo}. 
 }
\label{tab:main results}
\end{table*}

\section{Testbed: \dataset}
\label{section:dataset}
To better analyze the understanding misalignment problem of LLMs, we propose the benchmark \textbf{\dataset}, which contains challenge sentences that are prone to mistranslation.
This benchmark is constructed by collecting samples that multiple SOTA systems translate poorly from multi-year WMT testsets of six language pairs (\textit{Chinese} (\texttt{zh}), \textit{Estonian} (\texttt{et}), and \textit{Icelandic} (\texttt{is}) to/from \textit{English} (\texttt{en})).
Additionally, we believe that this dataset could promote future research in understanding the limitations of existing MT systems.

We select three SOTA MT systems: Google Translate, ChatGPT, and NLLB~\cite{Costa-jussa2024}. 
Due to the poor performance of NLLB in the zh$\Leftrightarrow$en translation, we additionally train a zh$\leftrightarrow$en translation model based on DeltaLM~\cite{ma2021deltalm} on the parallel corpus from OPUS\footnote{\url{https://opus.nlpl.eu/}}.
Next, all of the system translations are scored with COMET metric, and the $\rho$ of samples with the lowest score for each system are extracted as its difficult samples set. 
We vary the value of $\rho$ across different language pairs to ensure an appropriate scale for each difficult sample set.
Finally, the intersection of all systems' difficult sample sets is taken as the \dataset~testbed. 
\dataset comprises around 600+ sentence pairs for each language pair, which is illustrated in Tab.\ref{tab:Detail Statistics of dataset}.
We equally split this dataset into the validation set and the test set.

We report the translation performance measured by COMET on \dataset~and the complete WMT set in Fig.\ref{fig:performance change}, which shows that the performance decreases dramatically on \dataset~(84.5$\Rightarrow$73.6 averagely). 
Next, we conduct a multi-aspect comparison for \dataset~and the complete set in Appendix~\ref{section:Detail Statistics of dataset}, and find that the samples of \dataset~have higher perplexity (214$\Rightarrow$252 averagely). This result indicates that \textbf{sentences in \dataset~are more complex}.

\section{Experiments}
\subsection{Experimental Setup}
\paragraph{Comparative Methods.}
We verify the effectiveness of our \textsc{\method} on the LLM \texttt{GPT-3.5-turbo} for its promising  capability in following complicated instructions.
Demonstrations of \textsc{\method} are gained by performing our automatic method~(\S\ref{subsection:Automatic Construction of Demonstrations}) on the validation set of \dataset.
We compare \textsc{\method} with the following methods:
\begin{itemize}[leftmargin=*]
    \item \textbf{Zero-shot}, which asks the LLM to translate the source sentence directly.
    \item \textbf{ICL} (In-Context Learning), enhancing the translation with $K$ randomly selected exemplars from the validation set.
    \item \textbf{CoT}~\cite{wei2022chain}, encouraging the LLM to resolve the problem step by step. In this work, we re-implement CoT by prompting the LLM to translate the source sentence step by step.
    \item  \textbf{MAPS}~\cite{he2023exploring}, incorporating the knowledge of \textit{keywords}, \textit{topic words}, and \textit{demonstrations similar} to the given source sentence to enhance the translation process, respectively. 
    \item Commercial and open-source systems. We also report the performance of \textbf{Google} Translate, \textbf{NLLB} (in zh$\Leftrightarrow$en translation, we replace NLLB with our trained MT model based on DeltaLM), and zero-shot translation based on \textbf{GPT4} (\texttt{GPT-4-turbo}).
\end{itemize}
For \textsc{\method} and other ICL-based methods, we select $K$=8 demonstrations (\ie 8-shot) to achieve a strong baseline performance.
More details of re-implementing the baselines under the few-shot setting are illustrated in Appendix~\ref{section:details of experiments}.

\paragraph{Metrics.} Following previous research of LLM-based MT~\cite{10.5555/3618408.3618846,chen2023iterative}, we adopt COMET~\cite{rei-etal-2020-comet} and BLEURT~\cite{sellam-etal-2020-bleurt} as the evaluation metrics as their high correlations with human judgment than BLEU~\cite{papineni-etal-2002-bleu}.

\subsection{Results on \dataset}
\begin{table}[t]
\renewcommand\tabcolsep{6.0pt} 
\renewcommand\arraystretch{1.0}
\small
\centering
\begin{tabular}{lcccc}
\toprule
\multirow{2}{*}{\textbf{Methods}} & \multicolumn{2}{c}{\textbf{En}$\Rightarrow$\textbf{Is}}  & \multicolumn{2}{c}{\textbf{Is}$\Rightarrow$\textbf{En}} \\
\cmidrule(lr){2-3}
\cmidrule(lr){4-5}
 & \scriptsize{COMET}   & \scriptsize{BLEURT} & \scriptsize{COMET}   & \scriptsize{BLEURT} \\
\midrule
\textbf{Zero-shot} & 77.33 & 61.15 & 79.39 & 68.40 \\

\textbf{ICL} & 80.1	 & 61.99 & 81.02 & \textbf{70.20} \\

\textbf{\method-E} & \textbf{81.7} & \textbf{64.01} & \textbf{81.21} & 69.22\\ 
\bottomrule
\end{tabular}
\caption{Results in En$\Leftrightarrow$Is translation based on GPT-4.}
\label{table:results of \method based on GPT-4}
\end{table}
The main results are illustrated in Tab.\ref{tab:main results}. From the results, we have drawn the following observations:

\paragraph{(1) \textsc{\method} achieves significant improvements.} 
On average, \textsc{\method-E} surpasses the baseline ICL by +1.82 COMET and +1.32 BLEURT, and improves Zero-shot by +3.41 COMET and +3.33 BLEURT.
In the low-resource translation of En$\Rightarrow$Is, \textsc{\method-E} improves ICL by +3.85 COMET and Zero-shot by +5.87 COMET.
These improvements show that, \textsc{\method} largely elicits the translation ability via aligning the translation-specific understanding to the general one.

\paragraph{(2) \textsc{\method} achieves state-of-the-art performance on \dataset.} 
\textsc{\method} achieves the highest scores in En$\Leftrightarrow$Zh and En$\Leftrightarrow$Et translation in terms of COMET and BLEURT.
In En$\Leftrightarrow$Is translation, GPT-4 achieves the best results due to its superior multilingual capabilities.
To verify the effectiveness of \textsc{\method} on LLMs with stronger multilingual capabilities, we implement \textsc{\method} based on GPT-4 in En$\Leftrightarrow$Is translation, as shown in Table~\ref{table:results of \method based on GPT-4}.
The results suggest that \method~can also benefit stronger multilingual LLMs by facilitating the understanding misalignment.

\begin{table}[t]
\renewcommand\tabcolsep{6.0pt} 
\renewcommand\arraystretch{1.0}
\small
\centering
\begin{tabular}{lcccc}
\toprule
\multirow{2}{*}{\textbf{Methods}} & \multicolumn{2}{c}{\textbf{En}$\Rightarrow$\textbf{Zh}}  & \multicolumn{2}{c}{\textbf{Zh}$\Rightarrow$\textbf{En}} \\
\cmidrule(lr){2-3}
\cmidrule(lr){4-5}
 & \scriptsize{COMET}   & \scriptsize{BLEURT} & \scriptsize{COMET}   & \scriptsize{BLEURT} \\
\midrule
\textbf{WMT22 Best} & 86.80 & -- & 81.00 & -- \\
\hdashline[4pt/5pt]
\noalign{\vskip 2pt}
\textbf{Zero-shot} & 86.91 & 72.51 & 82.55 & 71.12 \\

\textbf{ICL} & 87.10 & \textbf{73.15} & 82.71 & 71.38 \\

\textbf{\method-E} & \textbf{87.60} & 72.41 & \textbf{82.75} & \textbf{71.75}\\ 
\bottomrule
\end{tabular}
\caption{Results on the complete WMT2022 testset. The result of WMT22 Best is reported for comparison.}
\label{table:results on the complete WMT.}
\end{table}

\paragraph{(3) CoT works poorly in machine translation.} In Table~\ref{tab:main results}, CoT incurs a dramatic performance drop over the baseline ICL.
Our case studies reveal that CoT produces extremely wordy translations, which is also observed by \citet{peng-etal-2023-towards}. 
We conjecture that CoT makes LLMs imitate junior translators rather than advanced translators.

\paragraph{(4) Difficult words are the bottleneck in translating complex sentences.}
The results show that incorporating the analysis of keywords and topics~\cite{he2023exploring} has yet to gain significant improvements as \textsc{\method}.
It suggests that it is the difficult words that lead to the performance bottleneck in translating intricate sentences.
We also follow \citet{he2023exploring} to experiment under the rerank setting as shown in Appendix.\ref{subsection:resuls under the rerank setting}, which shows the effectiveness of our method further.

\paragraph{(5) The external tool helps LLMs better identify difficult words.} 
\method-I achieves an average improvement of +1.52 COMET, demonstrating the effective performance of LLMs in recognizing difficult words.
And \method-E gains a further improvement of +0.3 COMET, showing the effectiveness of the external token-level QE tool in this task.

\subsection{Results on the complete WMT}
Experimental results on \dataset~show the effectiveness of \method~in translating complex sentences, raising the question of its impact on translating simple sentences.
Therefore, we conduct additional experiments on the complete WMT2022 testset of Zh$\Leftrightarrow$En translation.
As the results shown in Tab.\ref{table:results on the complete WMT.}, our method achieves comparable results to the baseline ICL in both translation directions.
These results show that \method~has no negative impact on translating simple sentences.

\begin{table*}[t]
\renewcommand\tabcolsep{4pt} 
\renewcommand\arraystretch{1.0}
\small
\centering
\begin{tabular}{lcccccc}
\toprule
& \textbf{En$\Rightarrow$Zh} & \textbf{Zh$\Rightarrow$En} & \textbf{En$\Rightarrow$Et} & \textbf{Et$\Rightarrow$En} & \textbf{En$\Rightarrow$Is} & \textbf{Is$\Rightarrow$En} \\

\midrule

\textbf{\#Mistranslation} & 19 & 25 & 26 & \textbf{53} & 22 & 34 \\
\textbf{\#Generalization Failure of Baseline} & 5 (26\%)  & 4 (16\%)  & 7 (27\%)  & \textbf{17 (32\%)} & 5 (23\%)  & 10 (29\%)  \\ 
\hdashline[4pt/5pt]
\noalign{\vskip 2pt}
\textbf{Translation Literalness of Baseline} & 4.11 & 3.53 & \textbf{4.75} & 4.46 & 3.76 & 3.70 \\
\bottomrule
\end{tabular}
\caption{Human evaluation results. }
\label{table:human evaluation results}
\end{table*}

\section{Analysis}
\subsection{Human Evaluation} \label{subsection:human evaluation}
To quantitatively analyze the understanding misalignment problem, we employ one senior human translator for each language direction to assess generalization failures and translation literalness.
Specifically, in each direction, we randomly sample 100 sentences\footnote{It takes 1.4 dollars for annotating one sentence.} from \dataset~and ask the senior translator to annotate the mistranslated words and phrases (\ie mistranslation) and score the literalness (1 to 5 score) of the translations generated by the strong baseline (\texttt{GPT-3.5-turbo} with ICL). 
Next, we ask the LLM to generate interpretations of the mistranslated content. These interpretations are presented to the translator, who judges whether they contain the correct understanding of the content. If the interpretation is accurate, the content is annotated as a generalization failure.
Finally, the translator is asked to annotate the generalization failures of \method and score the literalness of the translations produced by \method.

\paragraph{Analysis of generalization failures.} 
As the results shown in Tab.\ref{table:human evaluation results}, generalization failures account for a considerable proportion of all mistranslations (16\%$\sim$32\%).
Our method (\method-E) largely resolves these failures by 71\%$\sim$88\%.
We further study the unresolved failures and find that most of these unresolved failures are because they are not identified as difficult words by the LLMs in the stage of difficult word detection (Eq.\ref{eq:\method-E}) despite actually hard to translate.
It indicates that the difficult word detection remains an open question. 

\paragraph{Analysis of translation literalness.}
As the results shown in Tab.\ref{table:human evaluation results}, the baseline translations are highly biased towards literal translation.
\method significantly reduces the bias towards literal translation, indicating that the process of interpreting the difficult words first and then translating aligns better with sense-for-sense translation.

\subsection{Ablation Study}
\begin{table}[t]
\renewcommand\tabcolsep{6.0pt} 
\renewcommand\arraystretch{1.0}
\small
\centering
\begin{tabular}{lcccc}
\toprule
\multirow{2}{*}{\textbf{Methods}} & \multicolumn{2}{c}{\textbf{En}$\Rightarrow$\textbf{Zh}}  & \multicolumn{2}{c}{\textbf{Zh}$\Rightarrow$\textbf{En}} \\
\cmidrule(lr){2-3}
\cmidrule(lr){4-5}
 & COMET & $\Delta$ & COMET & $\Delta$ \\
\midrule
\method-E & 77.57 & -- & 73.23 & -- \\ 
\quad w/o. Draft & 76.94 & -0.63 & 72.68 & -0.55 \\
\quad w/o. IQC & 76.54 & -1.03 & 72.91 & -0.32 \\

\midrule
\midrule
\method-I & 76.92 & -- & 72.94 & -- \\ 
\quad w/o. Draft & 76.68 & -0.24 & 72.78 & -0.16 \\
\quad w/o. IQC & 76.45 & -0.47 & 72.59 & -0.35 \\
\bottomrule
\end{tabular}
\caption{Ablation Study. $\Delta$ indicates the performance drop after removing the specific component.}
\label{table:ablation study}
\end{table}
\textsc{\method} introduces the processes of (1) \textit{\textbf{draft translation}} to precisely detect the difficult words and (2) \textit{\textbf{IQC}} to improves the helpfulness of interpretations.
To clearly elucidate the contribution of these two components, we conduct an ablation study in Table~\ref{table:ablation study}.
Specifically, we analyze the effect of the draft translation by asking the LLM to detect difficult words directly without the draft translation.
The impact of IQC is analyzed by evaluating the performance of the generated translations guided by the original noisy interpretations (\ie without the processing of IQC).
The results show that removing either component leads to performance drops, and IQC plays a more important role in \textsc{\method}.
Specifically, the improvement of \textsc{\method} is halved when ablating the IQC on the En$\Rightarrow$Zh translation.

\subsection{Analysis of Difficult Word Detection}
\begin{figure}[t]
  \centering
    \includegraphics[clip,width=1.0\columnwidth,]{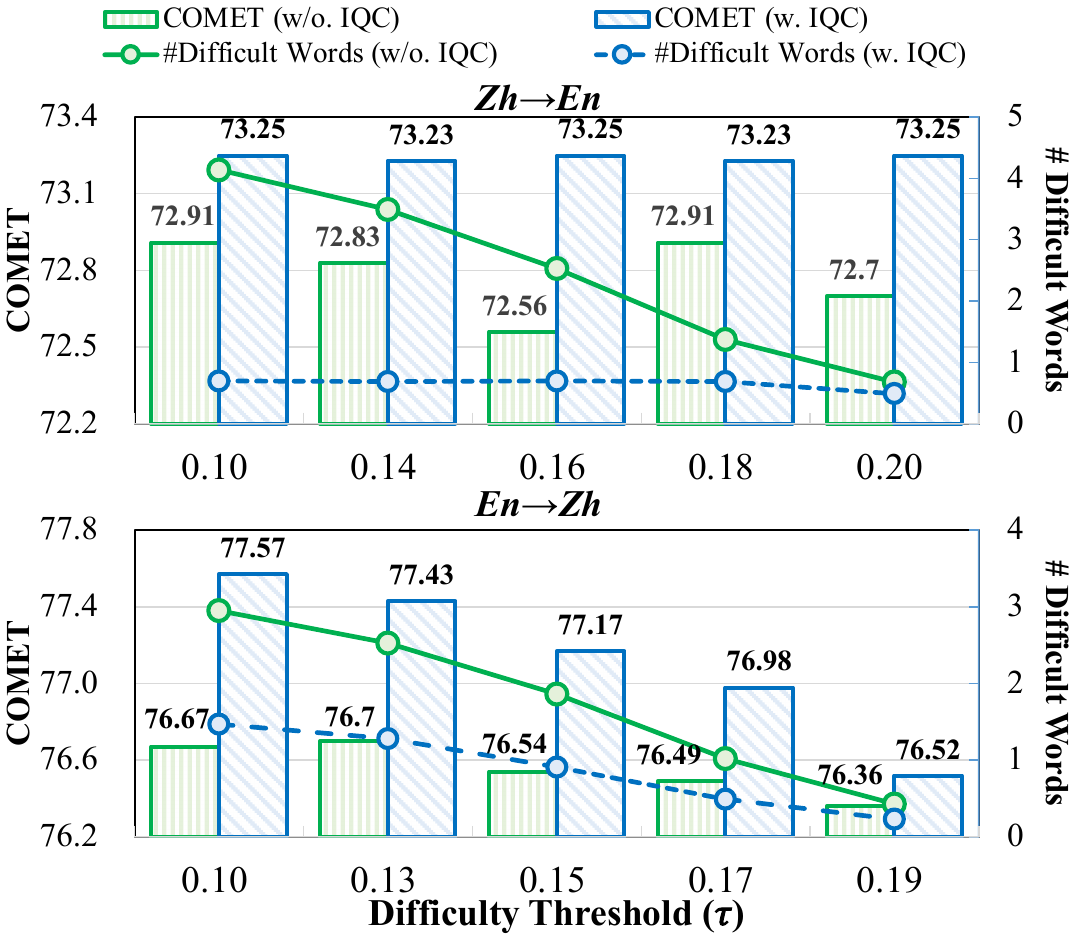}
    \caption{Effect of different values of difficulty threshold ($\tau$) on \textsc{\method-E}.}
  \label{fig:Effect of Difficulties NUmber}
\end{figure}

To offer an in-depth insight into the process of difficult word detection, we illustrate the relation between the number of difficult words interpreted and the resulting performance by adjusting the value of the difficulty threshold ($\tau$), which is shown in Fig.\ref{fig:Effect of Difficulties NUmber}.
Concretely, a smaller value of $\tau$ allows more difficult words to be interpreted.
From the results, we have the following observations:
\paragraph{Increasing the number of interpretations does not necessarily lead to performance improvements, but increasing high-quality ones can.}
Specifically, without controlling the quality of the interpretations (\ie w/o. IQC), increasing the number of interpretations (the \textcolor{Green2}{\textbf{green lines}}) yields unpredictable performance changes (as shown by the \textcolor{Green2}{\textbf{green bins}}), as introducing either valuable information or noise.
Fortunately, with IQC filtering negative interpretations, increasing the number of interpretations (the \textcolor{Blue2}{\textbf{blue lines}}) leads to constant improvements (as the \textcolor{Blue2}{\textbf{blue bins}} show).

\paragraph{Interpreting words that are more difficult brings larger improvements.}
Specifically, in the En$\Rightarrow$Zh translation, decreasing the value of $\tau$ from $0.19$ to $0.17$, the average number of helpful interpretations is increased from $0.23$ to $0.49$ ($+0.26$), and the performance is increased from $76.52$ to $76.98$ ($+0.46$).
However, decreasing the value of $\tau$ from $0.15$ to $0.10$, the average number of helpful interpretations is increased from $0.91$ to $1.47$ ($+0.56$), and the performance is increased from $77.17$ to $77.57$ ($+0.40$).
It should be noted that interpreting more words incurs more inference costs. Therefore, a modest value of $\tau$ (\ie $0.13 \sim 0.15$) is recommended to reach a compromise between efficiency and performance of \textsc{\method}.

\subsection{Analysis of Interpretation Generation}
\begin{figure}[t]
  \centering
    \includegraphics[clip,width=1.0\columnwidth,]{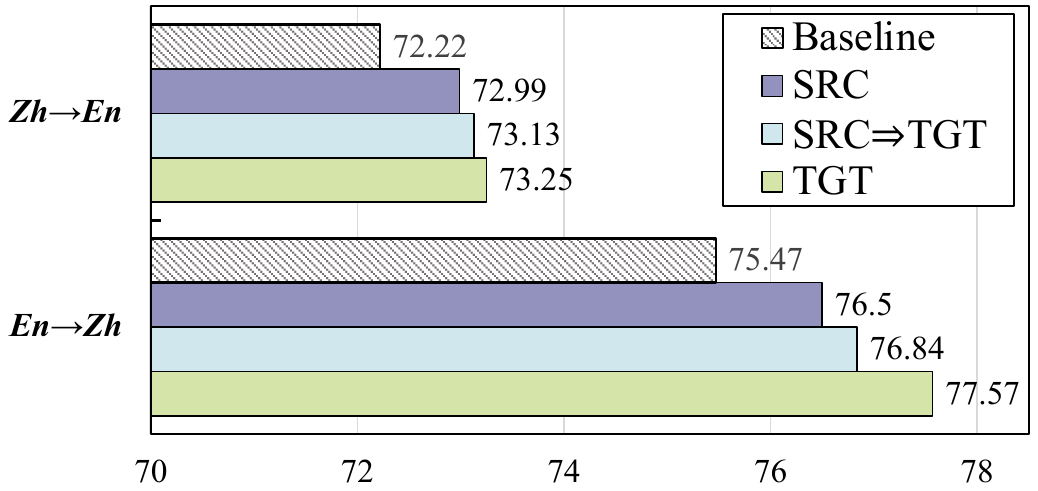}
    \caption{Effect of interpretations' language for \textsc{\method-E}.}
  \label{fig:Effect of interpretations' language}
\end{figure}

\paragraph{Languages of interpretations.}
Given a difficult word, \textsc{\method} generates the corresponding interpretation with the target language (\ie cross-lingual interpretation), which implicitly comprises two stages: (1) generating the interpretation in the source language and (2) translating the interpretation into the target language.
Compared with conducting these two stages explicitly, \textsc{\method} is more efficient and avoids error accumulation, which is illustrated in Fig.\ref{fig:Effect of interpretations' language}.
As demonstrated, interpretations in the target language (the \textcolor{Blue2}{\textbf{blue}} bins) are more beneficial than the ones in the source language (the \textcolor{Purple3}{\textbf{purple}} bins) owing to aligning the general understanding into the target language space, which could provide more benefits for translation.
And the implicit two-stage process (the \textcolor{Blue2}{\textbf{blue}} bins) is better than the explicit one (the \textcolor{Green2}{\textbf{green}} bins).

\section{Related Work}

\paragraph{Evaluation of LLMs' translation capabilities.}
With the remarkable progress of LLMs, researchers have assessed their translation abilities in various aspects. 
\citet{10.5555/3618408.3620130,vilar-etal-2023-prompting,10.5555/3618408.3618846,bawden-yvon-2023-investigating} first investigate LLM-based MT in terms of the prompt template and examples selection.
Next, the evaluation is extended across more domains~\cite{hendy2023good}, more languages~\cite{zhu2023multilingual}, and document-level translation~\cite{hendy2023good,wang-etal-2023-document-level}.
Other lines of work have performed in-depth assessments on the important attributes beyond accuracy, like literalness~\cite{raunak-etal-2023-gpts} and culture awareness~\cite{yao2023empowering}.
As existing studies have shown that LLMs have achieved promising performance, our work turns out to benchmark them on hard instances towards detecting more underlying issues.


\paragraph{LLM-based translation strategies.}
\citet{lu2023chainofdictionary} obtain the multilingual translations of keywords in the source sentence via the translator NLLB to augment the LLM, which improves the translation of low-resource languages while hurting the performance of high-source languages.
\citet{chen2023iterative} demonstrate that iterative refinement reduces translationese significantly.
\citet{he2023exploring} incorporate the knowledge of keywords, topics, and reference demonstrations to enhance the translation process, and use a rerank strategy to combine all candidate translations. 
However, there is no significant improvement to be observed when solely utilizing each single type of knowledge.
Different from previous works that utilize the intrinsic knowledge of LLM, \method~focuses on dealing with the difficult-to-translate words instead of the keywords for the reason that we argue the difficult-to-translate words lead to the performance bottleneck due to the long-tail distribution of knowledge.

\paragraph{LLM-based Automatic Post-Editing (APE).} APE corrects the errors in the generated translation, aiming to bias the translation towards the distribution of the target language~\citep{chen2023iterative,koneru-etal-2024-contextual}. Differently, our work aims to leverage the powerful understanding abilities of LLMs to correct the misunderstanding of complicated concepts in the source sentence. Our work is in parallel with APE.

\section{Conclusion}
In this work, we propose a novel translation process, \textsc{\method}, to take the first step in resolving the misalignment between the translation-specific understanding and the general understanding.
Furthermore, we utilize the token-level QE tool to enhance the detection of difficult words and the sentence-level QE tool to remove harmful interpretations.
Human evaluation results on high-resource and low-resource language pairs indicate that \method~significantly facilitates the understanding alignment, which improves the translation quality and alleviates translation literalness.  

\section{Limitations}
Even though \textsc{\method} elicits the translation abilities of LLMs via unleashing the general understanding (intrinsic knowledge) of LLMs, they still struggle to translate concepts that require the incorporation of extrinsic knowledge, such as the translation of \textit{neologisms}.
However, Our approach lays the foundation for researching when and how to incorporate external knowledge.
Besides, \method~requires to prompt the LLM for several times, leading to an increase in latency.
This latency is mainly caused by our interpretation quality control (IQC) strategy, which sequentially ablates each generated interpretation.
Concretely, if $|\mathcal{D}|$ difficult words are identified, IQC needs to prompt the LLM for $|\mathcal{D}|$ times.

\section*{Acknowledgements}
Bing Qin is the corresponding author of this work. We thank the anonymous reviewers for their insightful comments. This work was supported by the National Natural Science Foundation of China (NSFC) (U22B2059, grant62276078), the Key R\&D Program of Heilongjiang via grant 2022ZX01A32, the International Cooperation Project of PCL, PCL2022D01, the Nature Scientific Foundation of Heilongjiang Province(YQ2021F006), and the Fundamental Research Funds for the Central Universities (Grant No.HIT.OCEF.2023018).

\bibliography{anthology,custom}
\clearpage

\appendix

\begin{figure*}[t]
  \centering
    \includegraphics[clip,width=2.0\columnwidth,]{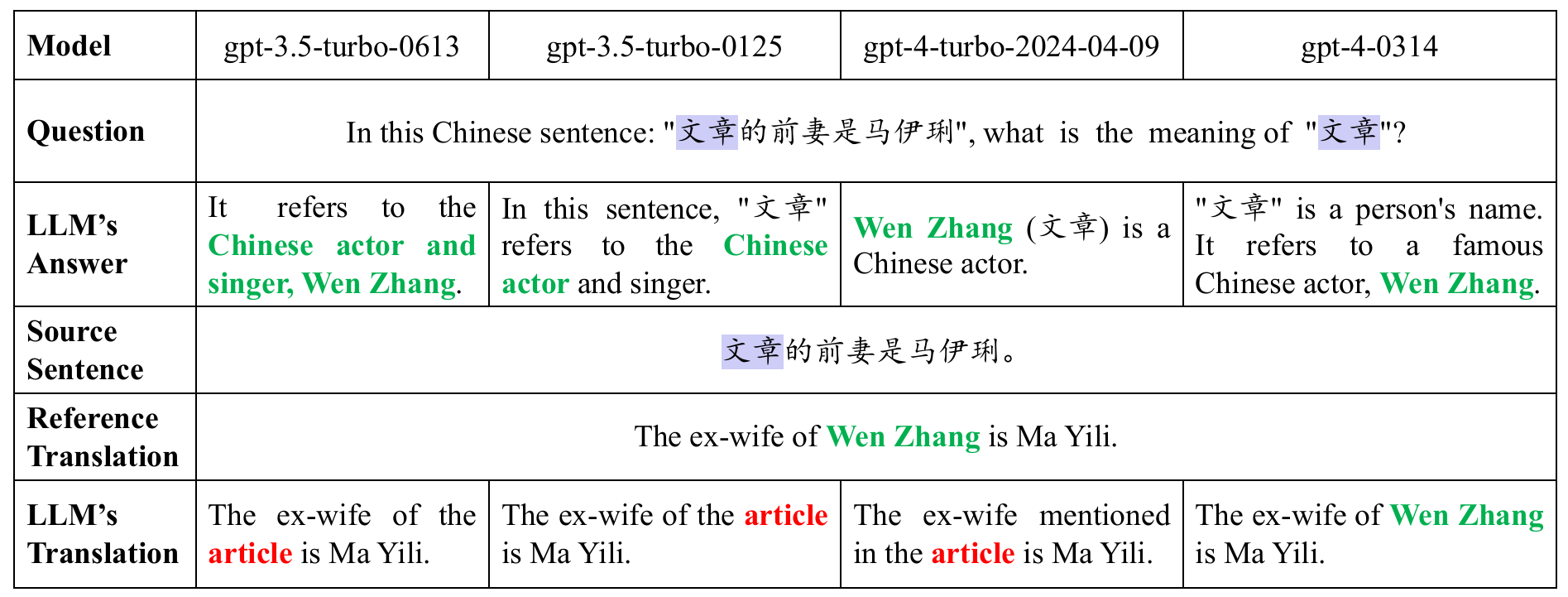}
    \caption{Illustration of understanding misalignment in more LLMs.}
  \label{fig:illustration of understanding misalignment in more LLMs}
\end{figure*}

\begin{figure*}[t] 
    \centering
    \begin{minipage}{0.45\linewidth}
        \centering
        \includegraphics[width=\linewidth]{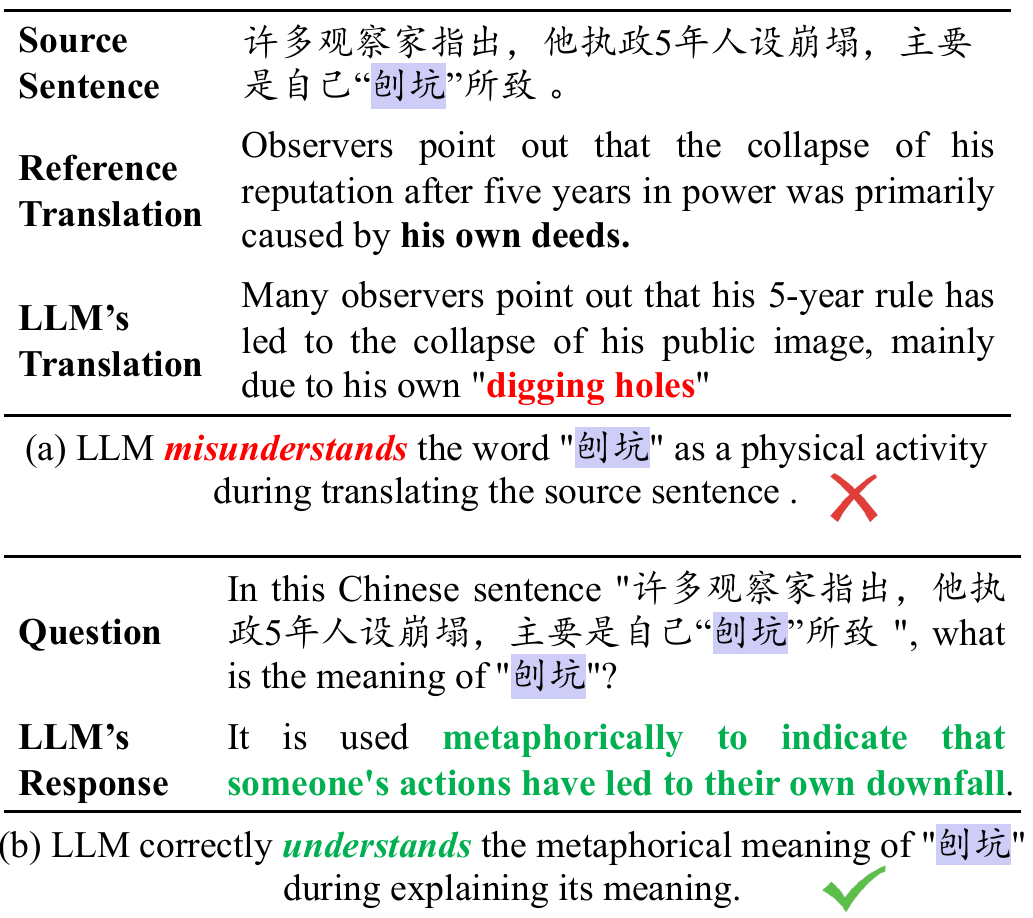}
        \caption{Illustration that understanding misalignment leads to LLMs literally translating some complicated concepts.}
        \label{fig:more_example_1}
    \end{minipage}
    \hspace{0.02\linewidth} 
    \begin{minipage}{0.45\linewidth}
        \centering
        \includegraphics[width=\linewidth]{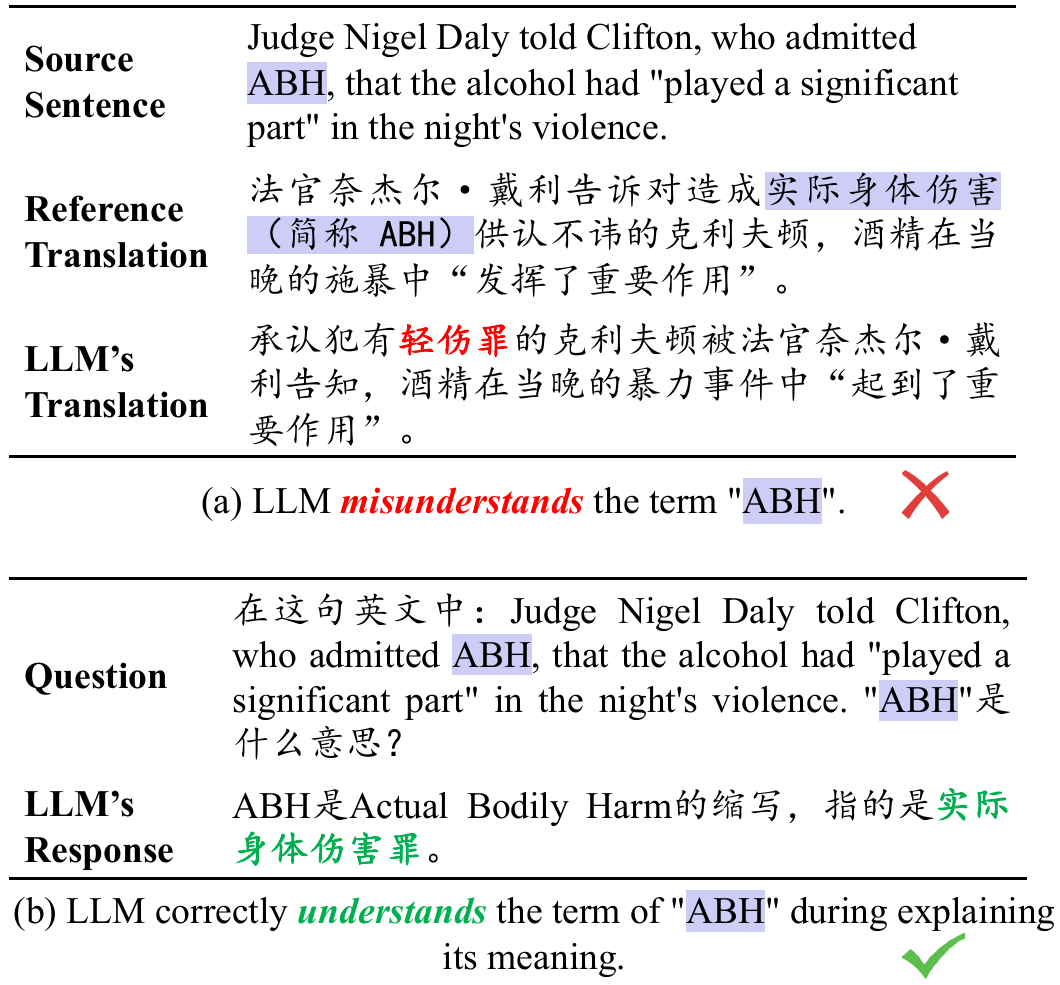}
        \caption{Illustration that understanding misalignment leads to LLMs mistranslating some terminology.}
        \label{fig:more_example_2}
    \end{minipage}
\end{figure*}



\begin{figure*}[t]
  \centering
    \includegraphics[clip,width=2.0\columnwidth,]{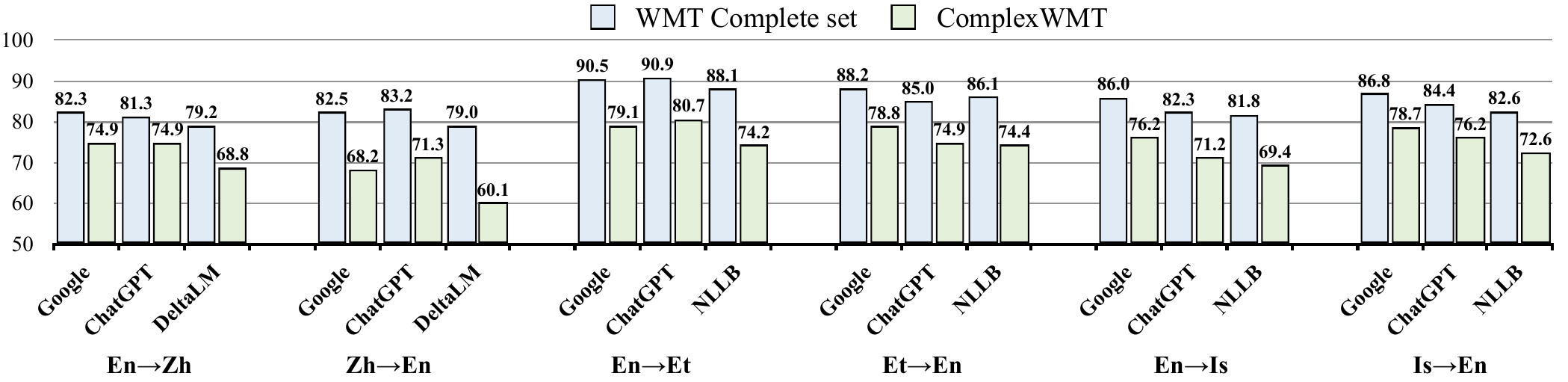}
    \caption{Translation performance on the complete WMT test set and the \dataset~test set.}
  \label{fig:performance change}
\end{figure*}

\begin{table*}[!t]
\small
\renewcommand\tabcolsep{3.0pt}
\renewcommand\arraystretch{1.15}
\centering
\begin{tabular}{ccccccccccccccc}

\toprule
Language pair & \multicolumn{2}{c}{\textbf{En$\Rightarrow$Zh}} & \multicolumn{2}{c}{\textbf{Zh$\Rightarrow$En}} & \multicolumn{2}{c}{\textbf{En$\Rightarrow$Et}} & \multicolumn{2}{c}{\textbf{Et$\Rightarrow$En}} & \multicolumn{2}{c}{\textbf{En$\Rightarrow$Is}} & \multicolumn{2}{c}{\textbf{Is$\Rightarrow$En}} & \multicolumn{2}{c}{\textbf{Average}}  \\

\cmidrule(lr){2-3}
\cmidrule(lr){4-5}
\cmidrule(lr){6-7}
\cmidrule(lr){8-9}
\cmidrule(lr){10-11}
\cmidrule(lr){12-13}
\cmidrule(lr){14-15}
Dataset & \textit{Comp.} & \textit{Chal.} & \textit{Comp.} & \textit{Chal.} & \textit{Comp.} & \textit{Chal.} & \textit{Comp.} & \textit{Chal.} & \textit{Comp.} & \textit{Chal.} & \textit{Comp.} & \textit{Chal.} & \textit{Comp.} & \textit{Chal.} \\

\midrule
\#Samples & 6215 & \textbf{675} & 7207 & \textbf{615} & 4000 & \textbf{644} & 4000 & \textbf{602} & 3004 & \textbf{641} & 3004 & \textbf{694} & 4572 & \textbf{645}\\
SRC-Len & 22.4 & \textbf{24.2} & 47.4 & \textbf{52.0} & 19.6 & \textbf{20.3} & 14.9 & \textbf{15.1} & 21.4 & \textbf{24.6} & 18.9 & \textbf{20.8} & 24.1 & \textbf{26.2}\\
TGT-Len & 42.5 & \textbf{50.9} & 28.9 & \textbf{34.1} & 14.9 & \textbf{15.5} & 19.6 & \textbf{20.8} & 20.6 & \textbf{25.1} & 20.4 & \textbf{23.2} & 24.5 & \textbf{28.3}\\
SRC-PPL & 141 & \textbf{165} & 40 & \textbf{79} & 128 & \textbf{156} & 823 & \textbf{925} & 111 & \textbf{147} & 40 & \textbf{40} & 214 & \textbf{252}\\
\#Noun & 4.2 & \textbf{4.9} & 3.4 & \textbf{4.1} & 4.6 & \textbf{4.7} & \textbf{1.7} & 1.6 & 5.8 & \textbf{7.1} & \textbf{2.1} & 2.0 & 3.6 & \textbf{4.1}\\
\#Verb & 5.2 & \textbf{5.9} & 3.1 & \textbf{3.7} & 3.0 & \textbf{3.2} & 2.5 & 2.5 & 3.8 & \textbf{4.4} & 2.6 & \textbf{2.8} & 3.4 & \textbf{3.8}\\
\#NE & 0.8 & \textbf{1.1} & 2.4 & 2.4 & \textbf{0.9} & 0.8 & 1.6 & 1.6 & 0.9 & \textbf{1.2} & \textbf{1.9} & 1.8 & 1.4 & \textbf{1.5}\\

\bottomrule
\end{tabular}
\caption{Fine-grained comparison of the complete WMT test set (\textit{Comp.}) and the \dataset~subset (\textit{Chal.}).
  `NE' is the abbreviation of "Named Entities".}
\label{tab:Detail Statistics of dataset}
\end{table*}

\section{Generalization Failures on Translation} \label{section:more examples}
In this section, we first provide the illustration of generalization failures on more LLMs, as shown in Fig.\ref{fig:illustration of understanding misalignment in more LLMs}.
As we can see, all of four LLMs accurately comprehend the complex concept, which three out of them mistranslate this concept.
Then, we also provide more examples of generalization failures, as shown in Fig.\ref{fig:more_example_1} and Fig.\ref{fig:more_example_2}.
\section{More details of \method}
\subsection{Details of IQC}
We gives a formal description of our interpretation quality control in Alg.\ref{alg:IQC}.

\subsection{Details of Demonstration Synthesis} \label{subsection:details of demonstration synthesis}
Inspired by the idea of Auto-CoT~\cite{zhang2023automatic}, we utilize LLM to generate the difficult words $\mathcal{D}$ and corresponding interpretations $\mathcal{A}$ based on the given bilingual sentence pair $(x, y)$:
\begin{mybox}
    \ Request: Given a $[L_s]$ sentence and its $[L_t]$ translation, please output the most difficult-to-translate words in the source sentence and concisely analyze the meaning of these words. \\ 
    The input-output format is: \\
    \\
    \textit{\# the format description is omitted.} \\
    Source Sentence: \ \  [\SourceSentence{Source Sentence $x$}] \\
    Target Translation:\;  [\DraftTranslation{Target Translation $y$}] 
\end{mybox}
Then, the response is parsed via regular expression to extract the difficult words $\mathcal{D}$ and interpretations $\mathcal{A}$. 
Next, we remove the noisy interpretations through a process similar to IQC (Alg.~\ref{alg:IQC}).
The only difference is that the QE metric is replaced with the reference-based COMET~\cite{rei-etal-2020-comet} due to the available access to the reference translation.
Finally, the generated difficult words $\mathcal{D}$ and interpretations $\mathcal{A}$ can be assembled with the source and target sentence $(x,y)$ as demonstrations for each step of \textsc{\method}.
\section{Statistics of \dataset}
\label{section:Detail Statistics of dataset}
We compare the complete WMT test set and the \dataset~subset in terms of the length of source sentences, the length of target sentences, the perplexity of source sentences, average number of nouns, verbs and named entities in the source sentence.
The statistics is shown in Table~\ref{tab:Detail Statistics of dataset}.

\section{Details of Experiments}
\label{section:details of experiments}
We conduct experiments under the few-shot setting.
To obtain the demonstrations of CoT, we ask the LLM to output the step-by-step translation process in a manner of post-explanation (\ie, given the source sentence and its translation, requesting the LLM to generate the intermediate process).
To obtain the ones of MAPS, we let the LLM to perform translation with the specific strategy on the validation set, and assemble the generated intermediate process (\eg keywords) and the reference translation as demonstrations.


\setlength{\textfloatsep}{10pt}
\begin{algorithm}[t]

\DontPrintSemicolon
\SetInd{0.5em}{0.7em} 
\SetKwInput{KwIn}{Input\quad}
\SetKwInput{KwOut}{Output\ }
\SetKwInput{KwInit}{Initialize}
\small
\caption{IQC}\label{alg:IQC}
\KwIn{source sentence $x$, draft translation $\widetilde{y}$, interpretations of difficult words $\mathcal{A}$, $\qquad$ QE scorer $\psi(\cdot)$} 

\KwOut{helpful interpretations $\hat{\mathcal{A}}$, \qquad \qquad \qquad final translation $\hat{y}$}

 $\hat{\mathcal{A}} \gets \mathcal{A}$ \;
 $\hat{y} \gets \mathop{argmax} P_{\theta}(\mathcal{E}^{igt}, x, \hat{y}, \mathcal{A})$ \;
 $\hat{s} \gets \psi (\hat{y}\ |\ x)$ \;
 
\For{$i \gets 1 \ to\  |\mathcal{A}|$}{
    $\overline{y} \gets \mathop{argmax} P_{\theta}(\mathcal{E}^{igt}, x, \hat{y}, \mathcal{A} - \{\mathcal{A}_i\})$, \;
    $\overline{s} \gets \psi(\overline{y}\ |\ x)$ \;
    
    \If{$\overline{s} > \hat{s}$}{
        $\mathcal{A} \gets \mathcal{A} - \{\mathcal{A}_i\}$, $\hat{y} \gets \overline{y}$, $\hat{s} \gets \overline{s}$ \;
    }
}

\end{algorithm}
\section{Results under the Rerank setting}
\label{subsection:resuls under the rerank setting}
We follow \citet{he2023exploring} to conduct experiments additionally under the rerank setting.
For the baseline ICL, we run for 4 times with different sets of demonstrations, which are sampled randomly with seeds $\{1,2,3,4\}$, and adopt QE to select the best candidate as the final translation.
For MAPS, the final translation is selected from the candidates generated by the three strategies (`+topic', `+Keywords', and `+SimDems') and ICL (seed=1).
For \textsc{\method}, we select the final translation from the results of \textsc{\method} and ICL (seed=1).
The results are shown in Table~\ref{tab:experimental results under the rerank setting}.

\begin{table*}[!t]
\small
\renewcommand\tabcolsep{1.5pt}
\renewcommand\arraystretch{1.0}
\centering
\begin{tabular}{lcccccccccccccc}

\toprule
\multirow{2}{*}{\textbf{Methods}} & \multicolumn{2}{c}{\textbf{En$\Rightarrow$Zh}} & \multicolumn{2}{c}{\textbf{Zh$\Rightarrow$En}} & \multicolumn{2}{c}{\textbf{En$\Rightarrow$Et}} & \multicolumn{2}{c}{\textbf{Et$\Rightarrow$En}} & \multicolumn{2}{c}{\textbf{En$\Rightarrow$Is}} & \multicolumn{2}{c}{\textbf{Is$\Rightarrow$En}} & \multicolumn{2}{c}{\textbf{Average}}  \\

\cmidrule(lr){2-3}
\cmidrule(lr){4-5}
\cmidrule(lr){6-7}
\cmidrule(lr){8-9}
\cmidrule(lr){10-11}
\cmidrule(lr){12-13}
\cmidrule(lr){14-15}
 & COMET   & QE   & COMET   & QE   & COMET   & QE   & COMET   & QE   & COMET   & QE   & COMET   & QE & COMET   & QE\\
\midrule
\multicolumn{15}{c}{\textbf{\textit{Baselines}}} \\
\textbf{ICL} & 76.79 & 3.94 & 72.67 & 0.43 & 82.10 & 9.37 & 79.98 & 7.44 & 73.42 & -1.21 & 78.88 & 4.80 & 77.31 & 4.13\\
\textbf{MAPS} & 77.24 & 4.56 & 73.17 & \textbf{1.70} & 83.05 & 10.57 & 80.12 & \textbf{8.28} & 75.67 & 2.61 & 78.47 & 5.13 & 77.95 & 5.48\\

\midrule
\multicolumn{15}{c}{\textbf{\textit{Ours}}} \\
\textbf{\method-I} & 77.36 & 4.37 & 73.30 & 1.08 & 83.06 & 10.39 & \textbf{80.22} & 7.96 & 76.88 & 3.12 & 78.93 & 5.29 & 78.29 & 5.37\\
\textbf{\method-E} & \textbf{77.78} & \textbf{5.04} & \textbf{73.36} & 0.88 & \textbf{83.21} & \textbf{10.93} & 80.10 & 8.06 & \textbf{77.39} & \textbf{3.97} & \textbf{79.22} & \textbf{5.31} & \textbf{78.51} & \textbf{5.70}\\

\bottomrule
\end{tabular}
\caption{Experimental results under the rerank setting.
 }
\label{tab:experimental results under the rerank setting}
\end{table*}
\end{CJK}
\end{document}